\documentclass{article}
\usepackage{spconf,amsmath,graphicx,hyperref}
\usepackage{amssymb}
\usepackage{multirow}

\title{MAN: Latent Diffusion Enhanced Multistage Anti-Noise Network for Efficient and High-Quality Low-Dose CT Image Denoising}
\name{Tangtangfang Fang\sthanks{These authors contributed equally to this work.}, Jingxi Hu\footnotemark[1], Xiangjian He\sthanks{Corresponding author.}, Jiaqi Yang}
\address{School of Computer Science, University of Nottingham Ningbo China, Ningbo, China.}
\begin{document}
%\ninept
%
\maketitle
\begin{abstract}
While diffusion models have set a new benchmark for quality in Low-Dose Computed Tomography (LDCT) denoising, their clinical adoption is critically hindered by extreme computational costs, with inference times often exceeding thousands of seconds per scan. To overcome this barrier, we introduce MAN, a Latent Diffusion Enhanced Multistage Anti-Noise Network for Efficient and High-Quality Low-Dose CT Image Denoising task. Our method operates in a compressed latent space via a perceptually-optimized autoencoder, enabling an attention-based conditional U-Net to perform the fast, deterministic conditional denoising diffusion process with drastically reduced overhead. On the LDCT and Projection dataset, our model achieves superior perceptual quality, surpassing CNN/GAN-based methods while rivaling the reconstruction fidelity of computationally heavy diffusion models like DDPM and Dn-Dp. Most critically, in the inference stage, our model is over 60x faster than representative pixel space diffusion denoisers, while remaining competitive on PSNR/SSIM scores. By bridging the gap between high fidelity and clinical viability, our work demonstrates a practical path forward for advanced generative models in medical imaging.
\end{abstract}
\section{Introduction}
\label{sec:intro}
The clinical adoption of Low-Dose Computed Tomography (LDCT) is essential for minimizing patient exposure to ionizing radiation\cite{waheed2022impact}. However, the inherent trade-off is a significant increase in image noise, which can degrade diagnostic quality by obscuring fine anatomical structures and pathological indicators\cite{ddm2}. While numerous denoising algorithms have been proposed, the challenge of achieving high quality image reconstruction without imposing prohibitive computational burdens remains a critical open problem in medical imaging.

Current deep learning approaches for LDCT denoising are divided by a sharp conflict between performance and speed. On one hand, fast methods employing Convolutional Neural Networks (CNNs)\cite{redcnn} or GANs\cite{wganvgg,dugan} often introduce over-smoothing artifacts\cite{ldctbenchmark}, failing to preserve the subtle textures vital for diagnosis. On the other hand, pixel-space diffusion models \cite{ddpm,dndp,yu2024pet} have set a new state-of-the-art in generative quality but are clinically infeasible due to their immense computational demands, which requiring thousands of iterative steps and prohibitively long inference times even on high-end hardware. This creates a distinct gap for a method that can bridge the performance of diffusion models with the practical speed of CNNs.

In this paper, we address this challenge by introducing MAN, a novel framework that adapts the latent diffusion model (LDM)\cite{ldm} paradigm for efficient and high-quality Low-Dose CT image denoising. By performing the diffusion and denoising process in a compressed latent space, MAN drastically reduces computational complexity while preserving the powerful generative capabilities of diffusion. 

Our main contributions are threefold: 1) We propose the LDM-based framework, accelerated with a DDIM-style quick sampler, specifically designed for LDCT denoising to explicitly balance reconstruction quality and inference speed; 2) We design a comprehensive architecture featuring a perceptually-optimized autoencoder and an attention-based U-Net to ensure the preservation of fine, clinically-relevant details; 3)We provide extensive experimental validation demonstrating that MAN achieves superior perceptual quality, outperforming both classic and recent deep learning methods, while being over 60x faster than comparable pixel-space diffusion models.

\section{Method}
\label{sec:method}
\begin{figure}[htb]
\centering
\includegraphics[width=\columnwidth]{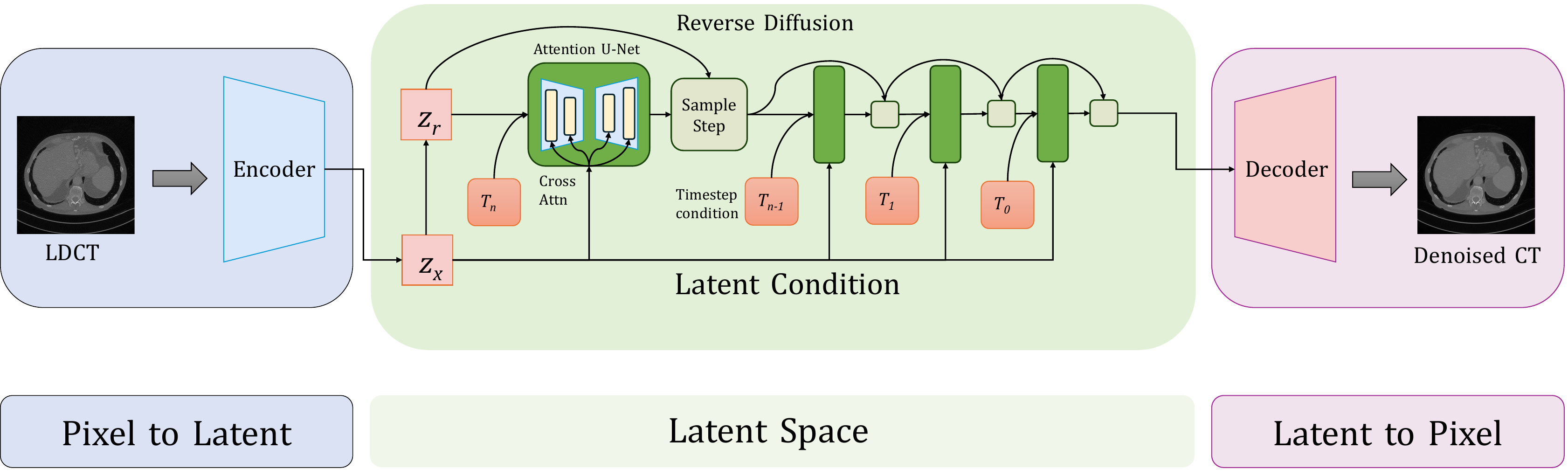}
\caption{Overview of the proposed MAN framework for LDCT denoising.}
\label{fig:framework}
\end{figure}

To achieve both high-quality denoising and fast inference, we propose MAN, a framework built on the latent diffusion model (LDM)\cite{ldm} paradigm. Our approach tackles two primary technical challenges: first, learning a compact yet faithful latent representation that preserves the fine anatomical details of CT images; and second, performing a computationally efficient yet powerful denoising process within this representation. We address these challenges with a two-stage approach, as illustrated in Figure \ref{fig:framework}: (1) a perceptually-optimized autoencoder, and (2) a conditional latent diffusion module accelerated with a DDIM-style\cite{ddim} quick sampler.

% \begin{figure*}[htb]
% \centering
% \includegraphics[width=\textwidth]{imgv4}
% \caption{Overview of the proposed MAN framework for LDCT denoising.}
% \label{fig:framework}
% \end{figure*}

\subsection{Perceptually-Optimized Autoencoder}
\label{ssec:ae}
The first stage is a customized autoencoder that maps a high-resolution input image $x \in \mathbb{R}^{H \times W \times 1}$ to a compressed latent representation $z = \mathcal{E}(x) \in \mathbb{R}^{h \times w \times c}$ and reconstructs it back to $\tilde{x} = \mathcal{D}(z)$. The goal is to create a perceptually rich latent space that is also regularized for the subsequent diffusion process.

The autoencoder is a U-Net-like architecture, it simplifies the autoencoder in the conventional LDMs\cite{ldm,khader2023denoising} based on the experience from PET-Diffusion\cite{petdif}'s approach, making it more suitable for the specific task of CT denoising. The encoder consists of stacked downsampling blocks, progressively reducing the spatial resolution while increasing the feature channels. Each block is composed of residual layers with Group Normalization and SiLU activation functions. To capture long-range dependencies, self-attention blocks are interleaved at the deep level. The deepest layer is a VAE-like\cite{vqvae} structure that maps the encoder's output to the latent space, outputting a latent vector. The decoder symmetrically mirrors this architecture using upsampling blocks to reconstruct the image.

The key to our autoencoder's effectiveness lies in its composite training objective, which ensures both pixel-accurate reconstruction and the learning of a smoothly regularized, perceptually-rich latent space. To this end, we combine four distinct loss functions: standard L1 and L2 pixel-wise losses, a KL-divergence loss $\mathcal{L}_{\text{KL}}$ from the original LDM's approach\cite{ldm}, and a perceptual loss $\mathcal{L}_{\text{PL}}$.

The KL-divergence loss regularizes the latent space by encouraging the posterior distribution of the encoder's output to be close to a standard normal distribution, which is a prerequisite for the subsequent diffusion process\cite{ldm}. It is defined as:

\begin{equation}
    \mathcal{L}_{\mathrm{KL}} = \frac{1}{2} \sum_{i=1}^{d} (\mu_i^2 + \sigma_i^2 - \log(\sigma_i^2) - 1).
\end{equation}

% $$ \mathcal{L}_{\text{KL}} = \frac{1}{2} \sum (\mu^2 + \sigma^2 - \log(\sigma^2) - 1) $$

The perceptual loss ensures that the reconstructions are visually realistic by penalizing differences in a deep feature space rather than in pixel space. We compute the L1 distance between feature maps extracted from a pre-trained VGG-16\cite{simonyan2014very} network at multiple layers ($l \in L$):

\begin{equation}
  \mathcal{L}_{\text{PL}} = \sum_{l \in L} || \phi_l(x) - \phi_l(\tilde{x}) ||_1.
\end{equation}

% $$ \mathcal{L}_{\text{PL}} = \sum_{l \in L} || \phi_l(x) - \phi_l(\tilde{x}) ||_1 $$
Although VGG16 model\cite{simonyan2014very} was trained on natural images, experiments provided by WGAN\cite{wganvgg} have demonstrated its reliability on medical images.

The final training objective is a weighted sum of these components. Crucially, we employ a dynamic scheduling strategy for the weights of the perceptual and KL losses. These terms are only introduced and gradually ramped up after an initial phase of pure reconstruction training. This strategy prevents the regularizing losses from interfering with the learning of basic features, leading to more stable training and a higher-fidelity final model. The total loss is:

\begin{equation}
  \mathcal{L}_{\text{AE}} = \mathcal{L}_{\text{pixel}} + \lambda_{\text{KL}}(t) \mathcal{L}_{\text{KL}} + \lambda_{\text{PL}}(t) \mathcal{L}_{\text{PL}},
\end{equation}

% $$ \mathcal{L}_{\text{AE}} = \mathcal{L}_{\text{pixel}} + \lambda_{\text{KL}}(t) \mathcal{L}_{\text{KL}} + \lambda_{\text{PL}}(t) \mathcal{L}_{\text{PL}} $$

where $\mathcal{L}_{\text{pixel}} = \lambda_1 \mathcal{L}_{\text{MSE}} + \lambda_2 \mathcal{L}_{\text{L1}}$ and $t$ represents the training step.

\subsection{Conditional Latent Diffusion}
\label{ssec:ldm}
The core denoising process is performed in the latent space using a conditional U-Net, $\epsilon_\theta(z_t, t, c)$, shown in Figure \ref{fig:framework}. This network is trained to predict the noise $\epsilon$ added to a clean latent vector $z_0$ at a given timestep $t$ and the conditioning context $c$ (derived from the input noisy image $x$). The U-Net architecture is composed of a series of residual blocks enhanced with spatial transformer blocks in its deeper layers. This allows the model to capture complex spatial relationships in the latent space. The timestep $t$ is encoded via sinusoidal position embeddings, and the conditioning information $c$ is injected into the U-Net through cross-attention mechanisms. This allows the denoising process to be guided by the content of the image being restored.

Instead of Linear schedule used in original model\cite{ddpm}, the forward diffusion process gradually adds Gaussian noise to the latent vector $z_0$ over $T$ steps according to a predefined cosine noise schedule\cite{nichol2021improved}, which defined as:
\begin{equation}
    \bar{\alpha}_t = \frac{\cos^2\left( \frac{(t / T + \epsilon)}{1 + \epsilon} \cdot \frac{\pi}{2} \right)}{\cos^2\left( \frac{\epsilon}{1 + \epsilon} \cdot \frac{\pi}{2} \right)}, \quad \text{with } \epsilon = 0.008,
\end{equation}
\begin{equation}
    \beta_t = 1 - \frac{\bar{\alpha}_{t}}{\bar{\alpha}_{t-1}}.
\end{equation}

For the reverse denoising process, we make a critical design choice to employ a deterministic DDIM-style\cite{ddim} quick sampler instead of the stochastic DDPM-style\cite{ddpm} sampler used in the original LDM\cite{ldm}. This significantly accelerates inference by reducing the required number of sampling steps to as few as 30 and ensures a stable, reproducible output for each input LDCT image. The reverse diffusion process iteratively refines the latent reconstruction from a noisy starting point $z_T$ back to a clean estimate $\tilde{z}_0$ using the following update rule:

% \begin{equation}
%     z_{t-1} = \sqrt{\bar{\alpha}_{t-1}}\left(\frac{z_t - \sqrt{1-\bar{\alpha}_t}\epsilon_\theta(z_t, t, c)}{\sqrt{\bar{\alpha}_t}}\right) + \sqrt{1-\bar{\alpha}_{t-1}}\epsilon_\theta(z_t, t, c).
% \end{equation}
\begin{equation}
\begin{aligned}
    z_{t-1} &= \sqrt{\bar{\alpha}_{t-1}}\left(\frac{z_t - \sqrt{1-\bar{\alpha}_t}\epsilon_\theta(z_t, t, c)}{\sqrt{\bar{\alpha}_t}}\right) \\
    &\quad + \sqrt{1-\bar{\alpha}_{t-1}}\epsilon_\theta(z_t, t, c).
\end{aligned}
\end{equation}

% $$ z_{t-1} = \sqrt{\bar{\alpha}_{t-1}}\left(\frac{z_t - \sqrt{1-\bar{\alpha}_t}\epsilon_\theta(z_t, t, c)}{\sqrt{\bar{\alpha}_t}}\right) + \sqrt{1-\bar{\alpha}_{t-1}}\epsilon_\theta(z_t, t, c) $$

Similar to conventional approaches\cite{ddpm,fastddpm}, the U-Net $\epsilon_\theta$ is trained with a simple mean squared error loss between the predicted noise and the ground truth noise:

\begin{equation}
  \mathcal{L}_{\text{LDM}} = \mathbb{E}_{z_0, \epsilon, t} \left[ || \epsilon - \epsilon_\theta(\sqrt{\bar{\alpha}_t}z_0 + \sqrt{1-\bar{\alpha}_t}\epsilon, t, c) ||^2 \right]
\end{equation}

The final denoised image $\tilde{x}$ is obtained by passing the refined latent vector $\tilde{z}_0$ through the autoencoder's decoder, $\tilde{x} = \mathcal{D}(\tilde{z}_0)$.

\section{Experiments}
\label{sec:exp}
\subsection{Experimental Setup}
\label{ssec:setup}
All experiments are conducted on a subset of the LDCT and Projection data archives\cite{mccollough2020low}, which is a widely used dataset for this task. Following established approaches\cite{redcnn,ldctbenchmark,dndp,fastddpm}, we use a subset of 1mm-slice abdominal and chest CT images (512x512) and use patient-wise split to create experimental dataset. We evaluate performance using Peak Signal-to-Noise Ratio (PSNR), Structural Similarity Index Measure (SSIM), and the Learned Perceptual Image Patch Similarity (LPIPS)\cite{zhang2018unreasonable}. LPIPS\cite{zhang2018unreasonable} is a perceptual metric that measures the similarity between two images based on deep features extracted from a pre-trained neural network. It captures perceptual differences that are more aligned with human visual perception. The LPIPS\cite{zhang2018unreasonable} is defined as:
\begin{equation}
    \text{L}(x, y) = \sum_{l} \frac{1}{H_l W_l} \sum_{h,w} \left\| w_l \odot \frac{\hat{\phi}_l(x)_{h,w} - \hat{\phi}_l(y)_{h,w}}{2} \right\|_2^2.
\end{equation}
% where $\hat{\phi}_l(x)_{h,w}$ and $\hat{\phi}_l(y)_{h,w}$ is the feature map extracted from the pre-trained network at layer $l$ and normalized to unit length, at spatial location $(h,w)$; $w_l$ is the learned layer weight; $H_l$ and $W_l$ are the height and width of the feature map; $\odot$ denotes element-wise multiplication.

Our proposed method, MAN, is benchmarked against a wide array of methods: the classic BM3D\cite{bm3d}; CNN-based RED-CNN\cite{redcnn} and QAE\cite{qae}; GAN-based WGAN-VGG\cite{wganvgg} and DU-GAN\cite{dugan}; and three representative pixel-space diffusion models, a standard DDPM\cite{ddpm} and the optimized Dn-Dp\cite{dndp} and Fast-DDPM\cite{fastddpm}.

\subsection{Implementation Details}
\label{ssec:impl}
Our method was implemented in PyTorch 2.6 with CUDA 12.6 and trained on a workstation equipped with two NVIDIA RTX 4080 Super GPUs. The autoencoder was pre-trained for 600 epochs with a gradient accumulation simulated batch size of 16. Similar to previous works\cite{ldm,fastddpm}, we used a dynamic strategy for the KL and perceptual loss weights, starting from 0 and increasing to their final values over the first 300 epochs. The pixel-wise loss weights were set to $\lambda_1=1$ and $\lambda_2=0.5$. The Adam optimizer was used with a learning rate of 1e-4. 

After the autoencoder training completed, its weights are fixed and used for the next stage of latent diffusion model training. As mentioned above, the conventional training target for a latent diffusion model is a U-Net-like noise prediction network, such as done in LDM\cite{ldm} and Brain Imaging Generation\cite{pinaya2022brain}. This study references these approaches by employing a DDIM-style\cite{ddim} diffusion control. Similar to PET-diffusion\cite{petdif}, the proposed latent diffusion model uses only Full-Dose CT (FDCT) images in its training, which simplifies the data acquisition requirements.

Both stages used the mixed precision training and data parallel training to maximize GPU memory usage.

\subsection{Quantitative Comparison}
\label{ssec:quant}
The main quantitative results are presented in Table \ref{tab:comparison}. Our proposed method, MAN, demonstrates a superior overall performance, excelling particularly in perceptual metrics. While RED-CNN\cite{redcnn} achieves the highest PSNR, a metric known to favor overly smoothed images\cite{ldctbenchmark,zhang2018unreasonable}, our method surpasses it and all other conventional methods in both SSIM and LPIPS. Notably, MAN achieves a LPIPS score of 0.1193, significantly outperforming all other conventional methods and indicating an advanced ability to reconstruct visually plausible images with fine textural details. Compared to the powerful pixel-space diffusion models DDPM\cite{ddpm} and Fast-DDPM\cite{fastddpm}, MAN achieves competitive reconstruction quality while offering orders of magnitude faster inference.

\begin{table}[htb]
    \centering
    \small
    \begin{tabular}{l|cccc}
        \hline
        \textbf{Method} & \textbf{PSNR} $\uparrow$ & \textbf{SSIM} $\uparrow$ & \textbf{LPIPS} $\downarrow$ & \textbf{Time} $\downarrow$\\
        \hline
        % LDCT & 26.14 & 0.527 & 0.236 \\
        BM3D\cite{bm3d} & 30.09 & 0.660 & 0.370 & 21.0m \\
        RED-CNN\cite{redcnn} & \textbf{31.38} & 0.693 & 0.272 & 3.7s \\
        Q-AE\cite{qae} & 30.19 & 0.652 & 0.246 & 0.6s \\
        WGAN-VGG\cite{wganvgg} & 27.03 & 0.533 & 0.300 & 0.3s \\
        DU-GAN\cite{dugan} & 30.35 & 0.683 & 0.211 & 3.8s \\
        DDPM\cite{ddpm} & 29.44 & 0.655 & 0.312 & 47m \\
        Fast-DDPM\cite{fastddpm} & 31.18 & 0.701 & 0.251 & 31.5s \\
        Dn-Dp\cite{dndp} & 29.02 & 0.623 & 0.211 & 18.7m \\
        \textbf{MAN (Proposed)} & 31.16 & \textbf{0.703} & \textbf{0.119} & 18.8s \\
        \hline
    \end{tabular}
    \caption{Quantitative comparison of different CT image denoising methods. Higher PSNR and SSIM values indicate better pixel level performance, while lower LPIPS values indicate better perceptual quality.}
    \label{tab:comparison}
\end{table}

The primary motivation for the proposed MAN is to achieve clinical viability. Table \ref{tab:comparison} includes the inference time for processing the entire test set. MAN requires only 18.8 seconds, which is not only comparable to fast CNN/GAN-based methods but represents a staggering ~60x speedup over pixel-space diffusion model like Dn-Dp\cite{dndp} and DDPM\cite{ddpm}. Compared to the sampling-optimized pixel-space diffusion method Fast-DDPM\cite{fastddpm}, the proposed MAN exhibits similar performance and slightly faster inference speed, which further demonstrates the potential power of combining latent diffusion\cite{ldm} with DDIM-style\cite{ddim} samplers. This result unequivocally demonstrates that our latent-space framework successfully bridges the gap between the generative power of diffusion models and the efficiency required for practical clinical deployment.

We acknowledge other important works that also aim to accelerate diffusion models, such as Fast-DDPM\cite{fastddpm}, which optimizes the sampling schedule, and ColdLDM\cite{coldldm}, which employs a cold diffusion framework. These methods report significant speedups for the original DDPM\cite{ddpm}. While their performance is impressive, our approach is fundamentally different and offers a complementary advantage. Instead of optimizing the sampling process of a computationally heavy pixel-space model, we drastically reduce the foundational computational load by shifting the entire diffusion process into a highly compressed latent space, and presenting a more holistic and scalable solution for high-resolution medical imaging tasks.

\subsection{Qualitative Results}
\label{ssec:qual}
\begin{figure}[htb]
\centering
\includegraphics[width=\columnwidth]{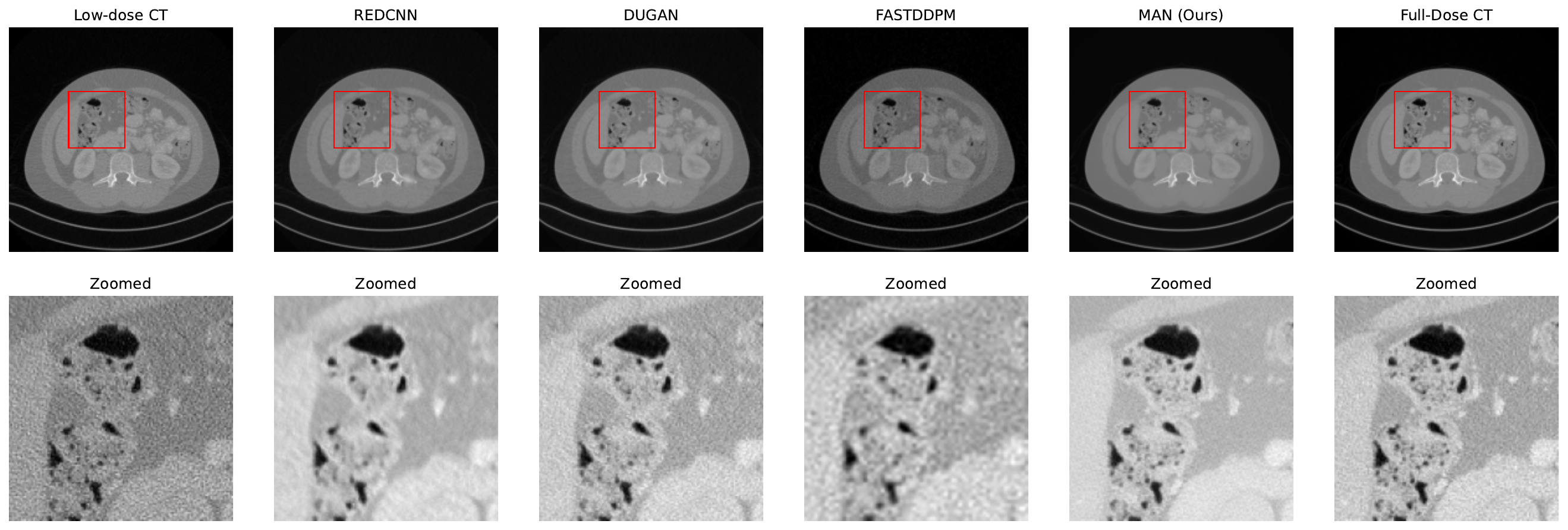}
\caption{Visual comparison of denoising results on a representative test slice. For fair visualization, the output of Fast-DDPM, originally 256x256, was upscaled to 512x512 using bicubic interpolation.}
\label{fig:visual}
\end{figure}

% Fig. \ref{fig:visual} provides a visual comparison of the denoising results on a representative test slice. The images show that while classic methods like RED-CNN\cite{redcnn} tend to produce blurry outputs that lose fine details, MAN successfully removes noise while preserving critical anatomical structures with high fidelity. Compared to other generative models like DU-GAN\cite{dugan}, our method yields images with fewer artifacts and a more natural texture, closely resembling the ground truth. This visual evidence confirms the superiority suggested by the LPIPS metric.
Figure \ref{fig:visual} provides a visual comparison of the denoising results on a representative test slice. The images clearly show the limitations of existing methods: RED-CNN\cite{redcnn} produces an overly smooth, blurry output that erases fine details; while DU-GAN\cite{dugan} and Fast-DDPM\cite{fastddpm} preserve textural features well but still exhibit residual noise. Our method, MAN, successfully avoids these pitfalls. It removes noise effectively while preserving critical anatomical structures with high fidelity. Compared to other methods, MAN yields images with a more natural appearance. This visual evidence confirms the superiority suggested by the LPIPS metric.

\subsection{Ablation Study}
\label{ssec:ablation}
To validate the effectiveness of our proposed components, we conducted a comprehensive ablation study, with the results summarized in Table \ref{tab:ablation}. We started with a baseline model, which is a standard latent diffusion\cite{ldm,petdif} model employing a basic autoencoder and a conventional DDPM-style sampler. Then we incrementally added our proposed components to isolate the impact of each design choice. We evaluated four distinct configurations: 1) A minimal LDM model using a simple autoencoder (trained with only pixel-wise L1/MSE and fixed KL losses) and a simple DDPM-style sampler; 2) The baseline with our perceptually-optimized autoencoder (P-AE); 3) The baseline with our DDIM-style quick sampler (Q-SP); 4) Our final proposed model, combining the proposed perceptually-optimized autoencoder (P-AE) with the proposed sampler (Q-SP).

\begin{table}[htb]
    \centering
    \small
    \begin{tabular}{cc|cccc}
        \hline
        \multicolumn{2}{c|}{\textbf{Components}} & \multicolumn{4}{c}{\textbf{Performance}} \\
        \hline
        \textbf{P-AE} & \textbf{Q-SP} & \textbf{PSNR} $\uparrow$ & \textbf{SSIM} $\uparrow$ & \textbf{LPIPS} $\downarrow$ & \textbf{Time} $\downarrow$\\
        \hline
        \texttimes & \texttimes & 29.72 & 0.676 & 0.156 & 3.29m\\
        \checkmark & \texttimes & 29.68 & 0.682 & 0.153 & 3.35m\\
        \texttimes & \checkmark & 30.93 & 0.695 & 0.132 & 18.98s\\
        \checkmark & \checkmark & \textbf{31.13} & \textbf{0.708} & \textbf{0.118} & \textbf{18.87s}\\
        \hline
    \end{tabular}
    \caption{Ablation study results.}
    \label{tab:ablation}
\end{table}

The results in Table \ref{tab:ablation} clearly demonstrate the individual and combined benefits of our contributions. While our P-AE module alone effectively improves perceptual metrics at a similar computational cost, our Q-SP provides the most dramatic improvement. It not only slashes the inference time by over 10-fold to under 19 seconds but also simultaneously enhances all quality scores, confirming the superiority of its deterministic path for this reconstruction task. Critically, our final model, combining both components, exhibits a strong synergistic effect, achieving the best overall performance and demonstrating that an optimized latent space and an efficient sampler are mutually beneficial.

% To validate our architectural design choices, we conducted an ablation study on the core components of our MAN model. We started with a baseline model and incrementally added our proposed contributions to isolate the impact of each design choice. We evaluated four distinct configurations: 1) A minimal LDM model using a simple autoencoder (trained with only pixel-wise L1/MSE and fixed KL losses) and a standard DDPM sampler; 2) The baseline with our perceptually-optimized autoencoder (P-AE) but still using the DDPM sampler; 3) Our final proposed model, combining the proposed autoencoder with the proposed sampler (Q-SP).

% The results, summarized in Table \ref{tab:ablation}, clearly demonstrate the individual and combined benefits of our design choices.

\section{Conclusions}
\label{sec:concl}
In this paper, we introduced MAN, the novel latent diffusion-based framework designed to address the critical trade-off between performance and speed in LDCT denoising. By operating in a compressed latent space with a perceptually-optimized autoencoder and a fast, deterministic quick sampler, our method achieves superior perceptual quality while remaining computationally efficient for clinical workflows. Our results demonstrate a significant improvement over a wide range of baselines, achieving an over 60x speedup compared to traditional pixel-space diffusion models without compromising reconstruction fidelity. While our model shows strong performance on the assumed noise distribution, future work could focus on extending this framework to explicitly model more complex, mixed noise types (e.g., Gaussian-Poisson), potentially further improving performance in diverse clinical scenarios.

% Below is an example of how to insert images. Delete the ``\vspace'' line,
% uncomment the preceding line ``\centerline...'' and replace ``imageX.ps''
% with a suitable PostScript file name.
% -------------------------------------------------------------------------

\vfill
\pagebreak

% \begingroup
% \setlength{\itemsep}{0pt}
% \setlength{\parskip}{0pt}
% \renewcommand{\baselinestretch}{0.95}\selectfont
% \bibliographystyle{IEEEbib}
% \bibliography{cite}
% \endgroup
\small
\bibliographystyle{IEEEbib}
\bibliography{cite}

\end{document}